\documentclass[conference]{IEEEtran}
\usepackage{algorithmic}
\usepackage[linesnumbered,ruled,vlined]{algorithm2e}
\usepackage{array}
\usepackage{verbatim}
\usepackage{amsthm}
\usepackage{amsmath,amssymb}
\usepackage{graphicx}
\usepackage{listings}
\usepackage{tikz}
\usepackage{color}
\usepackage{url}
\usepackage{wasysym}
\usepackage{multicol}
\definecolor{olivegreen}{rgb}{0.2,0.8,0.5}
\definecolor{grey}{rgb}{0.5,0.5,0.5}
\newtheorem{definition}{Definition}
\newtheorem{example}{Example}

\lstdefinelanguage{ttl}{
	sensitive=true,
	showspaces=false,
	showstringspaces=false,
	keywords={author, birthPlace, birthDate, broader,subject, label, completionDate, type, genre, philosophicalSchool,publicationDate, mainInterest, movement},
	morecomment=[l][\color{grey}]{--},
	morestring=[b][\color{blue}]\",
	morecomment=[s][\color{olivegreen}]{<}{>},
	alsoletter={-,<,>},
	emph={dbo,dbr,dc, Category, rdf, rdfs, skos}, emphstyle=\itshape,
	emph={[2]fun,cat, lincat,lin},emphstyle={[2]\color{red}}
}

\ifCLASSOPTIONcompsoc
\usepackage[caption=false,font=normalsize,labelfont=sf,textfont=sf]{subfig}
\else
\usepackage[caption=false,font=footnotesize]{subfig}
\fi

\begin{document}
	\title{Harmonization of conflicting medical opinions using argumentation protocols and textual entailment - a case study on Parkinson disease}

	% Fuzzy-based opinion reconcilation in Parkinson desease using argumentation protocols and textual entailment}

	\author{\IEEEauthorblockN{Adrian Groza and Madalina Mandy Nagy\IEEEauthorrefmark{1},
			\IEEEauthorblockA{\IEEEauthorrefmark{1}Intelligent Systems Group, \\
				Department of Computer Science,\\
				Technical University of Cluj-Napoca, Romania\\
				Adrian.Groza@cs.utcluj.ro, Mandy.Nagy@student.utcluj.ro
			}
		}
	}

	\maketitle
	
	\begin{abstract}
		Parkinson's disease is the second most common neurodegenerative disease, affecting more than 1.2 million people in Europe. 
		Medications are available for the management of its symptoms, but the exact cause of the disease is unknown and there is currently no cure on the market. 
		%In order to help organize the amount of existing information, we propose a system for resolving the conflicting situations using argumentation protocols and textual entailment. The knowledge base of the system is built from ontology and used in the textual entailment process as hypotheses. The aim is to create a multi-agent environment described by several conflicting opinions on common hypotheses. The mediation method combines textual entailment results and argumentation protocols. The uncertainty factor is controlled by fuzzy logic.
		To better understand the relations between new findings and current medical knowledge, we need tools able 
		to analyse published medical papers based on natural language processing and 
		tools capable to identify various relationships of new findings with the current medical knowledge.
		Our work aims to fill the above technological gap.
		%An example of these situations can be easily mapped on the data available on Web Environment. 
		%``To learn A, we did X''
		To identify conflicting information in medical documents, we enact textual entailment technology.
		To encapsulate existing medical knowledge, we rely on ontologies. 
		To connect the formal axioms in ontologies with natural text in medical articles, 
		we exploit ontology verbalisation techniques. 
		To assess the level of disagreement between human agents with respect to a medical issue, we rely on fuzzy aggregation. 
		To harmonize this disagreement, we design mediation protocols within a multi-agent framework. 
		%Our experimental results showed that .....
	\end{abstract}
	
	

	\section{Introduction}
	\label{sec:intro}
	Parkinson disease has still many unknowns, with weekly new findings that sometimes contradict each other. 
	An estimated seven to ten million people worldwide are living with Parkinson's disease.
	Parkinson's disease (PD) is a slowly progressive disorder of the brain that affects movement, muscle control, and balance. 
	PD manifests by loss of dopamine-producing cells in the brain, severely affecting the central nervous system. 
	Dopamine is a neurotransmitter whose decrease can affect a person's ability to control his movements, body and emotions. 
	%PD is a progressive illness, which means symptoms appear gradually and slowly get worse. 
	%Most people's symptoms take long time to develop, and they live for years with the disease before being diagnosed. 
	Ageing has been identified as an important risk factor; there is a two to four percent risk for Parkinson's among people over age 60, 
	compared with one to two percent in the general population.
	Despite decades of intensive study efforts, the causes of Parkinson's remain unknown. 
	It is unclear why exactly dopamine-producing cells become lost. 
	Many experts think that the disease is caused by a combination of genetic and environmental factors, which may vary for each person~\cite{environmentalFacts}. 
	
	Physicians have difficulties to cope with all newly published findings or clinical studies. 
	As medical experts interact with different medical studies, the divergence of opinions do exist in practice. 
	The conceptual unclarity about the cause of PD can lead to conflicting knowledge among medical experts 
	and several misunderstandings. 
	
	To better understand the relations between new findings and current medical knowledge, we need tools able 
	to analyse published medical papers based on natural language processing and 
	tools capable to identify various relationships of new findings with the current medical knowledge.
	Our work aims to fill the above technological gap.
	%An example of these situations can be easily mapped on the data available on Web Environment. 
	%``To learn A, we did X''
	To identify conflicting information in medical documents, we enact textual entailment technology.
	To encapsulate existing medical knowledge, we rely on ontologies. 
	To connect the formal axioms in ontologies with free natural text in medical articles, 
	we exploit ontology verbalisation techniques. 
	To assess the level of disagreement between human agents with respect to a medical issue, we rely on fuzzy aggregation. 
	To harmonize this disagreement, we design mediation protocols within a multi-agent framework. 
	%Nowadays, the approaching methods of conflicting information through textual entailment process are of high interest. 
	
	Our research question is how to disperse the conflicting information exchange on the subject of PD between two abstract agents. 
	The approach model is built on textual entailment results and information correctness combined in a mediation environment. 
	The knowledge base of the system is concentrated in hypotheses which are extracted from Parkinson's disease ontology. 
	The textual entailment method is used to train the system on a dataset constructed by these expert hypotheses and multiple text fragments available in the domain of PD. 
	The situation of conflicting information is then introduced in a multi-agent environment. 
	In cooperative knowledge-based systems decisions emerge from interrelationships and adaptation of the cognitive model of each agent involved~\cite{chaib2002causal}, \cite{Letia2006}, \cite{1565567}. 
	We plan to achieve a successful mediation process by correlating the argumentation protocols with the trained background knowledge. 
	The fuzzy logic components will help to design a more comparable reconciliation process to the real life situations, in which uncertainty is very often present.
	
	The remaining of the paper is organized as follows: 
	Section~\ref{sec:tech} introduces Textual Entailment, Description Logics, Fuzzy Reasoning and Argumentation Protocols as technical instrumentation used in the modeling process of the system. Section~\ref{sec:arch} presents the detailed architecture of our system and the tools which were used as it's components. 
	Section~\ref{sec:opinion} points out the steps involved for achieving opinion reconciliation. 
	Section~\ref{sec:scenario} presents the running scenario in which two agents are arguing on the correctness of a given hypothesis. 
	Section~\ref{sec:related} discusses related work, while section~\ref{sec:conclusion} concludes the paper.

	\section{Technical instrumentation}
	\label{sec:tech}
	
	The technical instrumentation used throughout the paper is based on three technologies:
	i) to identify conflicting information in medical documents, we enact
\emph{textual entailment} technology;
ii) to connect the formal axioms in ontologies with natural text in medical articles, we exploit
\emph{ontology verbalisation} techniques;
iii) to assess the level of disagreement between agents we use \emph{fuzzy reasoning}.

\subsection{Textual entailment}
	
	%\begin{definition}
		Textual entailment is a directional relationship between an entailing text $T$ and the entailed hypothesis,
		$h$, denoted by $T \models h$, if the meaning of $h$ can be inferred from the meaning of $T$.
	%\end{definition}
		Usually $h$ is a shorter sentence than the text $T$. 
	Recognizing textual entailment (RTE) task labels each entailment pair as
	either $Entailed$, $Contradicted$, or $Unknown$. A confidence factor is attached to each label. Consider the pairs $\langle T_1,h_1 \rangle$, $\langle T_2,h_2 \rangle$ and $\langle T_3,h_3 \rangle$ in the following examples:
	
	\begin{example}[Entailment $T \models h$] Let the pair $\langle T_1,h_1 \rangle$: 
		%\hspace*{-0.2cm}
		\begin{tabular}{lp{7.5cm}}
			$T_1$: & ``Hypomimia is a reduced degree of facial expression. It can be caused by motor impairment (for example, weakness or paralysis of the facial muscles), as in Parkinson's disease, or by other causes, such as psychological or psychiatric factors (for example, if a patient actually does not feel emotions and thus does not show any expression).''\\
			$h_1$: &``Hypomimia is a motor feature of Parkinson disease.''\\
		\end{tabular}
		\label{ex:entailment}
	\end{example}
	
	The information stated in $T_1$ explains the causes of a medical condition called \textit{hypomimia}.
	The hypothesis $h_1$ can be deduced by information extraction from $T_1$, noted by $T_1 \models h_1$. 
	The presence of motor impairments in the description of this medical condition in correlation with Parkinson's disease, labels hypomimia as a motor feature. 
	Therefore the pair is labeled with \textit{entailment}.
	
	\begin{example}[Contradiction $T\rightarrow\leftarrow h$] 
		\hspace*{-0.2cm}
		Let the pair $\langle T_2,h_2 \rangle$:
		
		\begin{tabular}{lp{7.5cm}}
			$T_2$: & ``Drugs used for diabetes and anti-malaria treatment have been suggested as disease modifiers in Parkinson's and perhaps even candidates to improve disease related symptoms. Preliminary data presented recently at the International Movement Disorders Society suggested that the diabetes and anti-malaria drugs were not effective in Parkinson's disease.''\\
			$h_2$: &  ``Malaria is an infectious cause of parkinsonism.''\\                
		\end{tabular}
		\label{ex:contradiction} 
	\end{example}
	
	The hypothesis $h_2$ states that malaria is considered a cause of Parkinson's disease, in contradiction with the information of $T_2$ which explains that medication for malaria has no effect on Parkinson's disease patients. This means that there could be no correlation between the two medical conditions. As a result, the pair is labeled with \textit{contradiction}.
	
	\begin{example}[Unknown $T\nleftrightarrow h$]  Let the pair $\langle T_3, h_3 \rangle$:
		%\hspace*{-0.2cm}
		\begin{tabular}{lp{7.5cm}}
			$T_3$:& ``Although doctors don't know exactly what causes Parkinson's disease, they think it is probably due to a combination of genetic and environmental factors.''\\
			$h_3$:& ``Cigarette smoking is an environmental risk factor for Parkinson's disease.''\\
		\end{tabular}
		\label{ex:unknown} 
	\end{example}
	
	The text fragment $T_3$ has no reference to the information offered by the hypothesis. $h_3$ is neither in contradiction or entailment with $T_3$. 
	This fact labels the pair with \textit{unknown}.
	
	For further details on models for recognizing textual entailment, the reader is referred to~\cite{dagan2013recognizing,sammons2010ask}.

	\subsection{Ontology verbalisation}
	
	%In the medical domain, the quality of information is essential. 
	%Description Logics (DLs) are the most expressive reasoning procedures, due to characteristics as complexity, scalability and connectivity. 
	The necessity of sharing common understanding of the structure of information among agents is the driven engine for developing ontologies~\cite{gruber}. 
Ontologies are formalised in Description Logics (DLs), characterized by a set of constructors provided for building complex concepts and roles. 
	These constructors are conjunction, disjunction, negation and restricted forms of quantification. 
	Table~\ref{table:dl1} presents the semantics of two concept descriptors $C$ and $D$, and also for the role $R$, given by interpretation $I = (\Delta ^{I}, \cdot ^{I}$), where $\Delta ^{I}$ stands for the domain of $I$ and $\cdot ^{I}$ is the interpretation function.

	The assertional part (ABox) is a set of assertions describing concrete situations, and the terminological part (TBox) is a set of axioms describing the structure of domain. 
	The Abox in Table~\ref{table:dl3} states that involuntary-shaking is an instance of the concept Tremor, while the individual Head-trauma is part of a medical risk factor. 
	In~\ref{table:dl4}), the concept Tremor is defined as motor feature that is part of parkinsonism, while the concept Motor-feature is included in those concepts that have at least one role part-of towards instances of Sign-of-parkinsonism.
	
	\begin{table*}
		\centering
		\caption{Semantics of DL.}
		\label{table:dl1}
		\begin{tabular}{||c | c | c | c ||} 
			\hline
			Constructor  & Syntax & Semantics & Examples \\ [0.5ex] 
			\hline
			atomic concept & A & $A^{I}\subseteq \Delta ^{I}$  & Tremor \\
			atomic role & R & $R^{I}\subseteq \Delta ^{I} X \Delta ^{I}$ & part-of \\
			conjunction & $C\sqcap D$ & $C^{I}\sqcap D^{I}$ & Tremor $\sqcap$ Primary-parkinsonism\\
			disjunction & $C\sqcup D$ & $C^{I}\sqcup D^{I}$ & Motor-features $\sqcup$ Nonmotor-features\\
			negation & $\neg C$ & $\Delta^{I} \setminus C$ &  $\neg$ Toxic-cause-of-parkinsonism\\
			existential restriction & $\exists R.C$ & $\left\{x | \exists y.\langle x,y\rangle \in  R^{I} \wedge y \in C^{I} \right\}$  & $\exists$is-a.Neuroprotective-agent-for-Parkinson \\
			value restriction & $\forall R.C$ & $\left\{x | \exists y.\langle x,y\rangle \in  R^{I} \Rightarrow  y \in C^{I} \right\}$  & $\forall$is-a.Models-of-Parkinson's-disease\\
			\hline
			
		\end{tabular}
	\end{table*}
	
	\begin{table}
		\centering
		\caption{ABox assertions.}
		\label{table:dl3}
		\begin{tabular}{|c|c|c|} 
			\hline
			Assertion & Syntax  & Example \\  
			\hline
			Concept  &$a:C$& involuntary-shaking:Tremor\\
			Role  &  $\langle a,b\rangle : R$& $\langle$ Head-trauma, Medical-risk-factors$\rangle$:part-of\\
			\hline
			
		\end{tabular}
	\end{table}

	\begin{table}
		\centering
		\caption{TBox axioms.}
		\label{table:dl4}
		\begin{tabular}{|c|c|c|} 
			\hline
			Axiom & Syntax & Example \\  
			\hline
			Definition   &  $A \equiv C$   & Tremor $\equiv$ Motor-feature $\sqcap$ $\exists$part-of.Parkinsonism\\
			Inclusion    & $C \sqsubseteq D$  &  Motor-feature $\sqsubseteq$ $\exists$part-of.Sign-of-parkinsonism \\
			\hline
			
		\end{tabular}
	\end{table}
	
	%We assume that the agents have a common ontology as starting knowledge base.   

	Natural Language Generation (NLG) aims at converting an input knowledge into an expression in natural language. 
	We are interested in NLG tools that produce textual summaries from ontologies, such a SWAT~\cite{stevens2011automating} %\footnote{http://swat.open.ac.uk/tools/}
	or NaturalOWL~\cite{galanis2007generating}. 
        %Paraphrasing is another method used for summarization, question answering and machine translation. 
        %To exemplify the concept, consider sentence $t_1$ and it's paraphrase $t_2$. $t_2$ will have the same or almost the same meaning as $t_1$. 
        %Compared to TE, paraphrasing can be seen as a mutual entailment.
	We used the NaturalOWL, an open-source natural language generation engine to obtain the hypotheses in Table~\ref{tab:verbalization}.

	%\begin{figure}
	%	\centering
	%\includegraphics[width=9cm]{ee.png}
	%	\caption{Example - part of ontology and resulted sentence}%
	%	\label{fig:f}
	%\end{figure}
	
	\begin{table*}
		\caption{Generating hypotheses from Parkinson ontology.}
		\centering
		\begin{tabular}{|l|l|l|}\hline
			Hypothesis & Axiom in DL & Axiom verbalisation \\ \hline
			$h_1$ & Hypomimia $\equiv$ Motor-feature  & Hypomimia is-a Motor-feature.\\ \hline 
			$h_2$ & Malaria $\equiv$ Infectious-cause-of-parkinsonism  & Malaria is-a Infectious-cause-of-parkinsonism.\\ \hline
			$h_3$ & Cigarette-smoking $\equiv$ Environmental-risk-factor  & Cigarette-smoking is-a Environmental-risk-factor. \\ \hline
			$h_4$ & Resting-tremor $\sqsubseteq$ Parkinsonian-tremor & Every Resting-tremor is-a Parkinsonian-tremor. \\ \hline
			$h_5$ & Canonical-etiology $\sqsubseteq$ Etiology-of-Parkinson-disease & Every Canonical-etiology part-of an Etiology-of-Parkinson-disease.
			\\ \hline
				\end{tabular}
		\label{tab:verbalization}
	\end{table*}

	\subsection{Fuzzy reasoning}
	In the fuzzy set theory~\cite{zadeh1999fuzzy} an element $x$ belongs to a given set $A$ with some degree expressed by the membership function $\mu_x(A)\Rightarrow [0,1]$. This membership function establishes the difference between crisp (classical) and fuzzy sets.
	
	%\begin{definition}
%		Let $\mathcal{A}$ be a set of agents. 
%		A fuzzy conflict (FC) is a binary fuzzy
%		relation $\mu_R:\mathcal{A} \times \mathcal{A} \Rightarrow [0,1]$.
%	\end{definition}

	Fuzzy knowledge is captured by fuzzy rules. 
         The Mamdani fuzzy rules difference is observed in the semantics of conventional operators and the fuzzy one's. The logical operators like $\wedge$ and $\vee$ are mapped on the T-norms and T-conorms such as min-max norm.
	
	%\begin{definition}
		Let V be a set of linguistic variables, such as strong, medium, weak. Fuzzification is a process of conversion data from the domain set to a set of linguistic variables such as V.
	%\end{definition}
	%\begin{definition}
		Defuzzification is the reverse of fuzzification. A common and useful defuzzification technique is center of gravity given by $z=\frac{\int_{Z}\mu_A(z)z\ dz}{\int_{Z}\mu_A(z)\ dz}$. The result of defuzzification is the crisp value $z$.
	%\end{definition}
	A fuzzy operation consists of three steps: 
		i) the membership function transforms the data into a fuzzy category by fuzzification;
		ii) all the rules are evaluated in order to obtain a fuzzy inference result
		iii) the result is transformed into a crisp value by defuzzification

	\section{System architecture}
	\label{sec:arch}
	
	\begin{figure*}
		\centering
		\includegraphics[width=16cm]{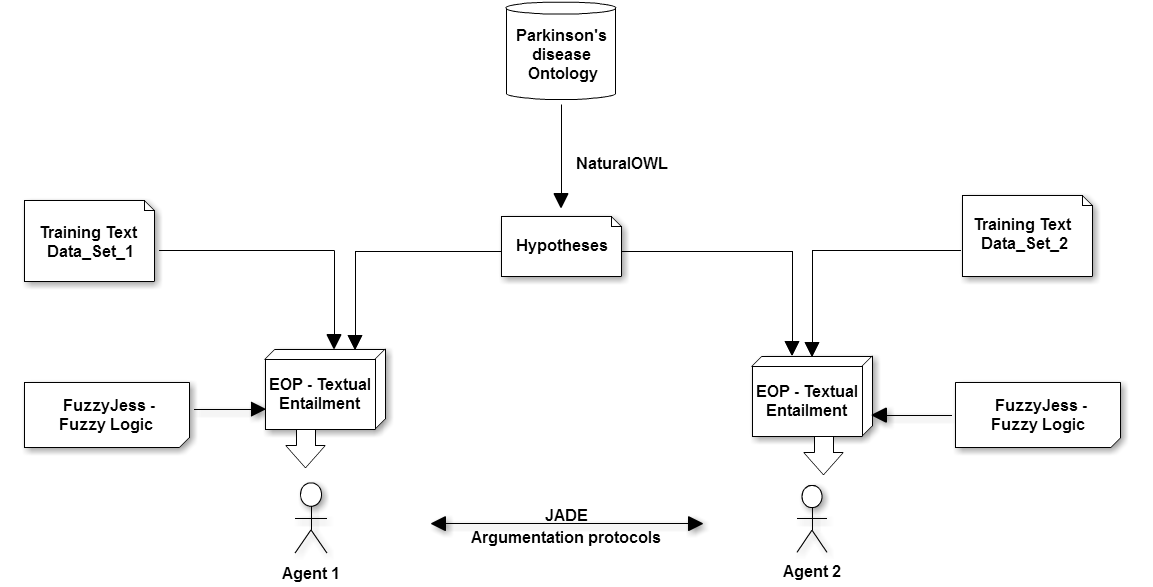}
		\caption{System architecture.}
		\label{fig:arch}
	\end{figure*}
	The developed system contains several modules depicted in Fig.~\ref{fig:arch}. 
		The first module is built around the NaturalOWL tool~\cite{galanis2007generating} and it is responsible to generate the set of hypotheses from the Parkinson's Disease Ontology (PDON). 
	The PDON ontology encapsulates the domain knowledge. 
	NaturalOWL is an open-source natural language generation system. %It's main function is related to the transformation of information from OWL format into coerent sentences. 
	NaturalOWL adopts a pipeline architecture, structuring the process into document planning, micro-planning and surface realization. 
	In the document planning step, the engine selects the relevant information from the ontology and groups those data which refer to a particular target subject. 
	 The system applies then templates on each information selected. 
	 Finally, the resulted text is computed through surface realization.

	A module is responsible to manage the new information or practical experience of the medical experts. 
	A corpus of documents from online sources (scientific papers, forums) is created for each agent in the system. 
	
	The Textual Entailment module is built on top of the Excitement Open Platform (EOP)~\cite{magnini2014excitement}. 
	EOP is a generic architecture and a comprehensive implementation for textual inference in multiple languages. %It's architecture consists of Linguistic Analysis Pipeline(LAP) and Entailment Core. 
       Three entailment decision algorithms (EDAs) are provided based on transformations, classifications and edit distance techniques. 
	
	 The multi-agent module is based on Jade platform~\cite{bellifemine2001jade}. 
	 The agents combine the textual entailment results with fuzzy logic and argumentation protocols. 
	 Hence, agents are capable to  detect, analyze, measure and resolve several conflicts among them. 
	 We assume that the agents share the same domain ontologies, but they have different training corpora encapsulating their private knowledge. 
	 We developed three protocols for conflict mediation. 
	The protocols rely on five possible locutions: informative statement, primary reason question, persuasive answer with data sharing, challenge, final remarks \{win, quit, lost\}. 
	At the end, the protocols force agents to declare their decision after the mediation process. 
	
%JADE is a distributed agents platform, implemented in Java. It provides an environment where JADE agents are executed, with class libraries to create agents using heritage and redefinition of behaviors and a graphical toolkit to help the management of agents.
	
The degree of agreement or conflict between agents is assessed by fuzzy reasoning. 
%The relational uncertainty factor inspired by the real world situations, are mapped on the system architecture by the integration of fuzzy logic elements. 
The corresponding knowledge is encapsulated as fuzzy rules within the FuzzyJess environment~\cite{friedman2003jess}. %FuzzyJess is a toolkit which can be used separately for creation and management of fuzzy rules, but it is often used in combination with Jess - a rule engine that uses an enhanced version of the Rete algorithm to process rules~\cite{friedman2003jess}. The fuzzy concepts are represented by fuzzy variables, fuzzy sets and values. 
Our agents recognize three types of fuzzy relations: \textit{agreement}, \textit{disagreement} and \textit{conflict}. 
We use the same fuzzy linguistic variables \textit{weak, medium, strong} to characterise the above fuzzy sets. 

The flow in which these modules are applied form the opinion reconciliation method described in the following section.

	\section{Opinion reconciliation}
	\label{sec:opinion}

Firstly, we present the top level algorithm of method for harmonizing opinions between agents. 
Then, we detail the fuzzy reasoning mechanism for assessing the degree of agreement, disagreement or conflict.
Finally, the mediation protocols for each situation are described.

	\subsection{Conflict mediation algorithm}
	
 The method for selecting the adequate protocol for current situation is formalised by Algorithm~\ref{alg:eval}. 
 The algorithm assumes that the domain ontology $\mathcal{O}$ and the available protocols $\mathcal{P}$ are common knowledge among agents. 
 We consider here three protocols: $P_a$ in case of agreement, $P_d$ in case of disagreement, and $P_c$ in case of conflicting opinions. 
 Algorithm~\ref{alg:eval} also assumes that the agents follow the same reasoning for situation awareness, by exploiting the same fuzzy knowledge base $\mathcal{FKB}$. 
 Hence, both agents identify that there is agreement or conflict on the given hypothesis. 
 That is, one agent does not consider a situation as a conflictual one, while its partner treats the same situation as an agreement. 
 The agents do have private knowledge representing their own experiences or readings. 
 This private knowledge is encapsulated by the corpus of medical documents $C_i$ of each agent $i$. 
 As output, the algorithm selects the adequate protocol from the set of common protocols $\mathcal{P}$.
 
 The set of medical hypotheses $\mathcal{H}$ is generated by verbalising the Parkinson ontology $\mathcal{O}$ (line 4).
 For each hypothesis $h$, each agent $i$  checks if $h$ is \textit{entailed}, \textit{conflicted}, or \textit{unknown} with respect to its private knowledge $C_i$ (lines 6 and 7). 
 The textual entailment procedure attaches also a certainty degree to the decision. 
In line 9, fuzzy reasoning is applied to identify the situation as agreement, disagreement or conflict with respect to hypothesis $h$. 
The initial set $\mathcal{H}$ is partitioned into three sets: 
 the set of agreeing hypotheses $\mathcal{A}$, the set of disagreeing hypotheses $\mathcal{D}$ and the  set of conflicting hypotheses $\mathcal{C}$.
 A degree of agreement is computed following a defuzzification method. 
 The resulted crisp value determines the most adequate protocol to harmonize the agents views on hypothesis $h$.
 	
	\begin{algorithm}
		
		\label{alg:eval}
		\caption{Selecting the harmonization protocol for each medical hypothesis.}
		
		\KwIn{$\mathcal{O}$ - Parkinson ontology;\\
			
			%$\mathcal{C}_2$ - corpus of medical information for agent $a_2$;\\
			$\mathcal{FKB}$ - fuzzy knowledge base;\\
			%$\mathcal{DG}$ - degree set;\\
			$\mathcal{P}=\{P_a, P_d, P_c\}$ - set of mediation protocols;\\
                        $\mathcal{C}_i$ - corpus of medical information for agent $i$;\\
			\KwOut{$P_i$, mediation protocol}
                $\mathcal{A}  \leftarrow\emptyset$ - set of agreing hypotheses;\\
			$\mathcal{D} \leftarrow \emptyset$ - set of disagreing hypotheses;\\
			$\mathcal{C}  \leftarrow \emptyset$ - set of conflicting hypotheses;}		

		$\mathcal{H} \leftarrow NaturalOWL(\mathcal{O})$

		\ForEach{$h \in \mathcal{H}$}{
                                        $\langle r_1,certainty_1 \rangle \leftarrow TextualEntailment(\mathcal{C}_1,h)$
 
                $\langle r_2,certainty_2 \rangle \leftarrow TextualEntailment(\mathcal{C}_2,h)$ 

               with $r_i \in \{ent, contr, unkn\}$ and $certainty_i \in [0..1]$ 
               
                $sit(h) \leftarrow \mathcal{FKB}(\langle r_1,certainty_1 \rangle,\langle r_2,certainty_2 \rangle)$

			\If {$sit(h) \equiv Agreement$} {$\mathcal{A} \leftarrow \mathcal{A} \cup \{h\}$}

\If {$sit(h) \equiv Disagreement$} {$\mathcal{D} \leftarrow \mathcal{D} \cup \{h\}$}
\If {$sit(h) \equiv Conflict$} {$\mathcal{C} \leftarrow \mathcal{C} \cup \{h\}$}
				
			$\mathcal{DG} \leftarrow defuzzification\_method(\mathcal{FKB} \langle \mathcal{A},\mathcal{D},\mathcal{C} \rangle) $
			
			$P_i \leftarrow protocol\_selection(\mathcal{P}, \mathcal{DG})$
			
			%$\mathcal{FC}(P_i)$

		}

	\end{algorithm}
	
	\subsection{Quantifying fuzzy agreement or disagreement}
	We define the commutative relation $\oplus$ between two decisions of textual entailment for agent $i$ and $j$.
	We use the following shortcuts: 
	entailment ($ent$), contradiction ($con$), unknown ($unk$), agreement ($a$), disagreement ($d$), conflict ($c$).

	%\begin{multicols}{2}
	%	\begin{itemize}

	%	\end{itemize}
	%\end{multicols}
	
	\begin{definition}
		An agreement occurs when the decisions of textual entailment process of the two agents are equivalent.
		\begin{align}
		ent_i  \oplus  ent_j  = a \label{eq:a1}\\
		con_i  \oplus  con_j  = a \label{eq:a2}\\
		unk_i \oplus unk_j = a \label{eq:a3}
		\end{align}
	\end{definition}

	\begin{definition}
		A disagreement occurs when the decisions of textual entailment process of the two agents are one of the following combinations:
		\begin{align}
		ent_i  \oplus  unk_j  = d \label{eq:d1}\\
		unk_i \oplus ent_j = d \label{eq:d2}\\
		con_i  \oplus  unk_j  = d \label{eq:d3}\\
		unk_j \oplus con_i = d \label{eq:d4}
		\end{align}
	\end{definition}

	\begin{definition}
		A contradiction is considered when the decisions of textual entailment process of the two agents are entailment, respectively contradiction. 
		\begin{align}
		ent  \oplus  con  = con \oplus ent = c  \label{eq:c1}
		\end{align}
	\end{definition}

	\begin{figure*}
		\begin{center}
			\begin{tabular}{lll}
			$R_1:$ &If $Ent(a_1,h_i)$ is strong $\wedge$ $Ent(a_2,h_i)$ is medium &$\Rightarrow$ $Agreement(h_i, a_1, a_2)$ is medium\\
			$R_2:$ &If $Ent(a_1,h_i)$ is medium $\wedge$ $Ent(a_2,h_i)$ is medium &$\Rightarrow$ $Agreement(h_i, a_1, a_2)$ is strong\\
			$R_3:$ &If $Ent(a_1,h_i)$ is strong $\wedge$ $Ent(a_2,h_i)$ is weak &$\Rightarrow$  $Agreement(h_i, a_1, a_2)$ is weak\\
			$R_4:$ &If $Ent(a_1,h_i)$ is strong $\wedge$ $Unkw(a_2,h_i)$ is medium &$\Rightarrow$ $Disagreement(h_i, a_1, a_2)$ is medium\\
				$R_5:$ &If $Contr(a_1,h_i)$ is strong $\wedge$ $Unkw(a_2,h_i)$ is strong &$\Rightarrow$ $Disagreement(h_i, a_1, a_2)$ is strong\\
				$R_6:$ &If $Unkw(a_i,h_1)$ is weak $\wedge$ $Ent(a_2,h_i)$ is strong &$\Rightarrow$ $Disagreement(h_i, a_1, a_2)$ is weak\\
				$R_7:$ &If $Contr(a_1,h_i)$ is strong $\wedge$ $Ent(a_2,h_i)$ is strong & $\Rightarrow$ $Conflict(h_i, a_1, a_2)$ is strong\\ 
				$R_8:$ &If $Contr(a_1,h_i)$ is medium $\wedge$ $Ent(a_2,h_i)$ is weak & $\Rightarrow$ $Conflict(h_i, a_1, a_2)$ is weak\\
				$R_9:$ &If $Contr(a_1,h_i)$ is medium $\wedge$ $Ent(a_2,h_i)$ is strong & $\Rightarrow$ $Conflict(h_i, a_1, a_2)$ is medium\\
				$R_{10}:$ &If $Contr(a_1,h_i)$ is weak $\wedge$ $Ent(a_2,h_i)$ is strong & $\Rightarrow$ $Conflict(h_i, a_1, a_2)$ is weak\\
			\end{tabular}
		\end{center}
		\caption{Sample of fuzzy rules for assessing conflicting opinions for agents $a_1$ and $a_2$ with respect to hypothesis $h_i \in \mathcal{H}$.}
		\label{fig:FKB1}
	\end{figure*}
	
\begin{figure}
		\centering
		\includegraphics[width=9cm]{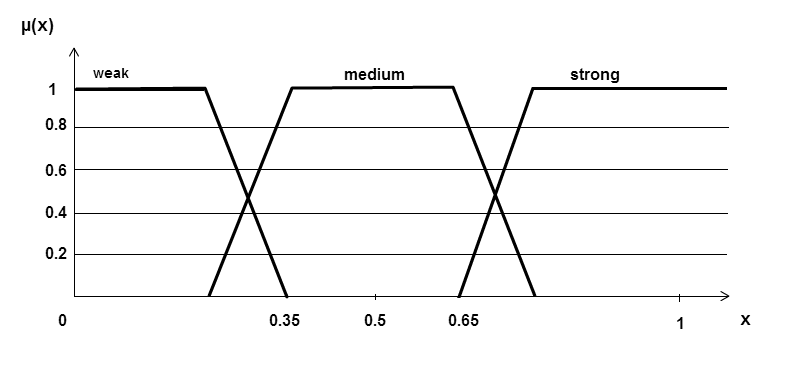}
		\caption{Fuzzy membership function of a the agreement relation.}
		\label{fig:FKB2}
	\end{figure}

	Let $a_1$ and $a_2$ be two agents.  
	Fuzzy conflicts allow expressing degree of agreement, disagreement or conflict among agents. 
	Assessing the degree of agreement, disagreement or conflict on each hypothesis $h_i$ 
	is computed with the fuzzy rules in Fig.~\ref{fig:FKB1}.
	The entailment ($ent$), contradiction ($contr$), and unknown ($unkn$) 
	relationships are defined based on three membership functions: strong, medium, and weak (see Fig.~\ref{fig:FKB2}). 
	Similarly, the agreement, disagreement and conflict are assessed based on the same membership functions.
	
	Given the set of hypotheses $\mathcal{H}$, based on fuzzy membership function, we can measure the degree of disagreement between two agents.
	The disagreement is a consequence of their learning experiences (train set) or learning method, that is in our case the textual entailment algorithm. 
	
	Let $|{A}|$, $|{D}|$, $|{C}|$, the cardinality of ${A}$, ${D}$ and ${C}$ relations for two agents $a_1$ and $a_2$.
		
		\begin{definition}[Cardinality-based metrics]
			 The similarity of opinions towards the set of hypotheses $\mathcal{H}$ is given by the agreement metric $agr(\mathcal{H})=\frac{|{A}|}{|H|}$.
			 The divergence of opinions towards the set of hypotheses $\mathcal{H}$ is given by the disagreement metric $dis(\mathcal{H})=\frac{|{D}|}{|H|}$.
			  The opposing opinion towards the set of hypotheses $\mathcal{H}$ is given by the conflicting metric $con(\mathcal{H})=\frac{|{C}|}{|H|}$.
				\end{definition}
		
	The following mediation protocols aim to reduce this disagreement. 
	
	\subsection{Designing mediation protocols for each situation}\label{sec:designing-argumentation-protocol-for-each-type-of-relation}
	
        Three mediation protocols are designed for each detected situation: $P_d$ in case of disagreement, $P_c$ in case of conflict, respectively $P_a$ in case of agreement between agents.

	\paragraph{Argumentation protocol for Disagreement ($P_d$)}
	\begin{figure}
		\centering
		\includegraphics[width=4.5cm]{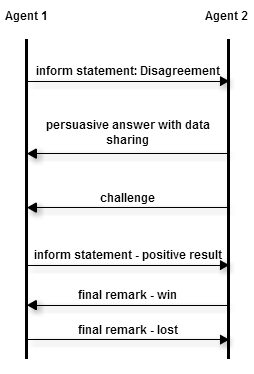}
		\caption{Enacting the disagreement protocol $P_d$.}
		\label{fig:disagree}
	\end{figure}
	The $P_d$ protocol is enacted when $d$ is deduced by one equation~(\ref{eq:d1}), (\ref{eq:d2}), (\ref{eq:d3}), or (\ref{eq:d4}). 
	Consider the situation $ent_i \oplus unk_j$.  
	A graphical example of disagreement protocol flow can be seen in Fig.~\ref{fig:disagree}.
	The agent $j$ seems to learn from  examples that are not related to the current hypothesis. 
	By providing relevant medical texts, agent $i$ can alter $j$' perspective on current hypothesis. 
	The quantity of exchanged information is decided by the providing agent. 
	One heuristic for the provider agent $i$ is to correlate the percentage of information shared with the degree of disagreement between them.
	The receiving agent $j$ is challenged to reconsider its answer after adding the received information to his textual entailment training. 
	The output of the entailment procedure can be altered or not, depending on the quantity and relevance of the examples exchanged. 
	If agent $j$ observes a positive change in his results, he reconsiders his opinion. 
	If not, he will challenge the other agent exactly as happened before. 
	After two challenges, the protocol ends and each agent makes a statement about his final opinion.
	
	\paragraph{Argumentation protocol for Conflict ($P_c$)}
	The $P_c$ protocol is enacted when $c$ is deduced by equation~(\ref{eq:c1}).
	Consider the situation $ent_i  \oplus  con_j  = con$.
	Here, the opinions are attacking each other, hence a strong disagreement.
	The protocol starts by sending an informative statement which declares the conflict.
	Then, the agent $i$ will take a primary reason question. 
	Due to the conflict situation between them, the protocol assumes that same number of training examples is shared at each step.
	This is differently from the protocol $P_d$, in which the agents decided the percentage individually. 
	The conflicting situation gives a degree of self-seeking for both agents. 
	The exchange of information will take place in three rounds, and at the end both $i$ and $j$ will declare their final opinion with respect to $h$.
	
	\paragraph{Argumentation protocol for Agreement ($P_a$)}
	The $P_a$ protocol is enacted when $a$ is deduced by equation~(\ref{eq:a1}), (\ref{eq:a2}) or (\ref{eq:a3}).
	In this case, the opinions of the two agents are more or less equivalent. 
	Consider the case $ent(i, h_1)=0.85$ and $ent(j, h_1)=0.45$. 
	Fuzzifying this crisp value, agent $i$ considers $h_1$ as strongly entailed with a degree of $0.93$.
	After accumulating knowledge from corpus $C_j$, agent $j$ considers hypothesis $h_1$ as medium entailed with a degree of $0.60$.
	Performing fuzzy inference in $\mathcal{FKB}$ with the min/max norms, the situation corresponds to a medium  agreement with degree $0.80$.
%	By performing defuzzyfication based on the center of mass method, the output crisp value is $0.71$.
	For this value, the following communicative acts may be conveyed: 
	This is a non-conflicting situation, in which, $i$ will make an informative statement about the relation:
	$inform(i, j, content(h_1, ent_i \oplus ent_j))$.
	Agent $j$ may question the degree of the result with $request\_explanation(j, i, FD(strong))$.  
	If it has different value than strong, the two agents will share a part of their best results. 
	If the degree has strong value, the dialog ends.
	
	By running the protocols $P_c$, $P_d$ and $P_a $ over each hypothesis $h$ we expect an increasing of the agreement over the set $\mathcal{H}$. 
	Considering a cost-benefit analysis of this mediation process, we define the following relation: %\begin{equation 
	$\Delta(level_x \rightarrow level_y) > \sum_{k=0}^{N}cost_k + \sum_{p=0}^{M}cost_p$ %\label{eq:costbenefit}
%\end{equation}
   Here, the left side of the equation presents the benefit of a mediation process expressed by the conversion of a certain level of agreement, disagreement or conflict to another. The right side of the equation defines the cost of a harmonization process. The first sum represents the cost of exchanged information, every training sample ($cost_k$) having a certain price. The second sum represents the cost of communicative acts quantified in terms of time ($cost_p$).

	\section{Running experiments}
	\label{sec:scenario}
	
%In order to illustrate the harmonization process achievements, we run a use case scenario showing each step from the initial state to the final results. 
Consider the agents $a_1$ and $a_2$ sharing the same Parkinson's Disease Ontology (PDON). 
The textual entailment hypotheses are generated through ontology verbalisation. 
For $a_1$, the training corpus $C_1$ contains 56 pairs of text-hypothesis (36\%\ Entailment, 33\%\ Unknown, 31\%\ Contradiction).
For $a_2$, the raining corpus $C_2$ contains 56 pairs of text-hypothesis (37\%\ Entailment, 31\%\ Unknown, 32\%\ Contradiction).
The test corpus $T$ is common for $a_1$ and $a_2$ and it contains 10 pairs of text-hypothesis.

We train the system using a training data set, and then test the performance on an unseen test dataset. During training, only the training set is available. The test set was locked during the training step. This stage of training and testing was made with two of the EDAs provided by the system and the highest accuracy was obtained with Classification-based EDA - 0.703\%. The text corpus of the training set contains 200 pairs of text-hypothesis and it was devided into 35\% of  entailment pairs, 32.5\% of contradiction pairs and 32.5\% of unknown pairs. The textual entailment algorithms return a confidence degree of every pair's decision. These decisions combined with confidence degrees will be used in the following layers. 
The results of Classification-based Entailment Decision Algorithm are presented in the third and forth column of Table~\ref{tabel:te}. 

\begin{table}
\caption{The sets $\mathcal{A}$, $\mathcal{D}$, $\mathcal{C}$ before and after running the mediation protocols.}
\begin{tabular}{|l|l|l|}\hline
Set & Before & After \\ \hline
$\mathcal{A}$ & $\{h_1,h_2,h_3,h_4,h_6,h_9,h_{10}\}$& $\{h_1,h_2,h_3,h_4,h_6,h_8, h_9,h_{10}\}$  \\
$\mathcal{D}$ & $\{h_7,h_8\}$ & $\{h_7\}$ \\
$\mathcal{C}$ & $\{h_5\}$ & $\{h_5\}$ \\ \hline
\end{tabular}
\label{tab:sets}
\end{table}

The sets $\mathcal{A}$, $\mathcal{D}$ and $\mathcal{C}$ contain the hypotheses illustrated in Table~\ref{tab:sets}:.
The initial agreement over the set $\mathcal{H}$ is quantified as $agr(\mathcal{H})=7/10$, the disagreement as $dis(\mathcal{H})=2/10$, while conflict as $con(\mathcal{H})=1/10$.
After running the corresponding mediation protocols for each hypothesis, the agreements are:  $agr(\mathcal{H})=8/10$, $dis(\mathcal{H})=1/10$, $con(\mathcal{H})=1/10$. 

\begin{table*}
			\centering
			\caption{Harmonizing medical knowledge between agents $a_1$ and $a_2$ on ten hypotheses.}
			\footnotesize\setlength{\tabcolsep}{2.5pt}
			\label{tabel:te}
			\begin{tabular}{|c|>{\centering}m{6cm}|c|c|c|c|c|}
				\hline
			& Verbalised hypothesis ($\mathcal{H}$)& Agent $a_1$  & Agent $a_2$ & Initial state & Selected & Final state\\  
			& & $\langle$TE, Confidence$\rangle$ & $\langle$TE, Confidence$\rangle$ & (before harmonization)& protocol& (after harmonization)\\
			\hline
			$h_1$ & Camptocormia is a Long term complication of medication. & $\langle$Unkw, 0.57$\rangle$ &  $\langle$Unkw, 0.50$\rangle$ & medium Agreement & $P_a$ & medium Agreement\\
				\hline
			$h_2$ &Malaria is an Infectious cause of parkinsonism. &  $\langle$Ent, 0.52$\rangle$ &  $\langle$Ent, 0.30$\rangle$ & weak Agreement & $P_a$ & medium Agreement\\
				\hline
			$h_3$ &Nocturia is a a Clinical marker associated with Parkinson's disease.& $\langle$Contr, 0.46$\rangle$ & $\langle$Contr, 0.56$\rangle$ & medium Agreement & $P_a$ & medium Agreement\\
				\hline
			$h_4$ &Tuberculosis is an Infectious cause of parkinsonism.& $\langle$Unkw, 0.54$\rangle$&$\langle$Unkw, 0.50$\rangle$ & medium Agreement & $P_a$ & medium Agreement\\
				\hline
			$h_5$ &Hypomimia is a Motor feature of Parkinson's disease.&$\langle$ Contr, 0.41$\rangle$&$\langle$ Ent, 0.68 $\rangle$ & medium Conflict & $P_c$ & weak Conflict\\
				\hline
		$h_6$ &Ocular surface irritation is part of Ophthalmological symptoms of Parkinson patients.&$\langle$Ent, 0.49$\rangle$&$\langle$Ent, 0.52$\rangle$ & medium Agreement & $P_a$ & medium Agreement\\
				\hline
			$h_7$ &Poliomyelitis is an Infectious cause of parkinsonism.&$\langle$Unkw, 0.54$\rangle$&$\langle$Contr, 0.45$\rangle$ & medium Disagreement & $P_d$ & medium Disagreement\\
				\hline
			$h_8$ &Hallucination is a Long term complication of medication.&$\langle$Ent, 0.50$\rangle$&$\langle$Unkw, 0.33$\rangle$ & weak Disagreement & $P_d$ & weak Agreement\\
				\hline
			$h_9$ &Inflammation Causes a Neuropathology of Parkinson's disease.&  $\langle$Unkw, 0.52$\rangle$&$\langle$Unkw, 0.50$\rangle$ & medium Agreement & $P_a$ & medium Agreement\\
				\hline
			$h_{10}$ &Mucuna pruriens is part of a Pharmacological management of Parkinson's disease.&$\langle$Contr, 0.47$\rangle$&$\langle$Contr, 0.54$\rangle$ & medium Agreement & $P_a$ & medium Agreement\\
				\hline
			%&&&& Agreements:  $\sfrac{7}{10}$ & &Agreements:  $\sfrac{8}{10}$\\
			%&&&& Disagreements:  $\sfrac{2}{10}$ && Disagreements:  $\sfrac{1}{10}$\\
			%&&&& Conflicts:  $\sfrac{1}{10}$ && Conflicts:  $\sfrac{1}{10}$\\
			%\hline
				
			\end{tabular}
	
		\end{table*}

In the following, we detail the harmonization process between $a_1$ and $a_2$ regarding the hypothesis 

$h_5$: "Hypomimia is a Motor feature of Parkinson's disease."
	
The system identifies the relation between $a_1$ and $a_2$. 
Firstly, the fuzzy engine places $a_1$ in the \textit{medium contradiction} category, and $a_2$ in the \textit{strong entailment} category. 
Secondly, the relation is determined by the fuzzy rule $R_9$ (recall Fig.~\ref{fig:FKB1}) applied on hypothesis $h_5$.  
Thirdly,  
	% \textit{If $Contr(a_1,h_5)$ is medium $\wedge$ $Ent(a_2,h_5)$ is strong $\Rightarrow$ $Conflict(h_5, a_1, a_2)$ is medium} 
the crisp value 0.46, places the result in \textit{medium conflict} situation. 
%We define three levels of conflict, mapping the fuzzy degrees \{strong, medium, weak\}:
%	 \begin{enumerate} 
%	 \item	$level_1$  $\equiv$  $Conflict(h_5, a_1, a_2)$ is weak
%	 \item 	$level_2$  $\equiv$ $Conflict(h_5, a_1, a_2)$ is medium
%	 \item	$level_3$  $\equiv$  $Conflict(h_5, a_1, a_2)$ is strong
%	\end{enumerate}
%	 The initial state is $i_{h_5}$ = $level_2$.
The harmonization procedure continues with the activation of the protocol $P_c$ assigned for conflict situations.
%described in Subsection~\ref{sec:designing-argumentation-protocol-for-each-type-of-relation}. 
%The protocol is activated in the Jade platoform
%in the multi-agent argumentation environment, see Figure~\ref{fig:sniffer}. 
The information exchange imposed by the protocol $P_c$ ends with the following declarations: $inform(a_2,a_1, WIN)$ while $a_1$ declares failure: $inform(a_1,a_2, LOST)$. 
The opinion of $a_1$ converts into \textit{weak contradiction} and the fuzzy rule $R_{10}$ infers a weak conflict.
Consequently, the mediation process managed to alter the strong conflict into a weak one.
	 %  The initial and final state of every hypothesis from Test Corpus $T$
 %is presented in the fifth and the last column of Table~\ref{tabel:te}. 

	%	\begin{figure}
	%		\centering
	%		\includegraphics[width=9cm]{sniff.png}
	%		\caption{Argumentation protocol for conflict - JADE - Sniffer Agent}
	%		\label{fig:sniffer}
	%	\end{figure}

	\section{Related Work and Discussion}
	\label{sec:related}
	%We approach related work from three perspectives: 
	Textual entailment (TE) is currently seen as a possible solution for a variety of natural language problems 
	such as question answering, summarization, and information extraction, machine translation~\cite{articolConflict}. 
	The combination of argumentation with textual entailment has been investigated in~\cite{cabrio2012combining}. 
	Here,  positive textual entailment represents support between arguments, while negative textual entailment represents attack between arguments.
	As the approach in~\cite{cabrio2012combining} relies on classical Dung's abstract argumentation framework, support relations have not been represented. 
	We rely on fuzzy reasoning to quantify the conflict between hypotheses and we design argumentation protocols to harmonize divergence of opinions.  
		%\paragraph{Argumentation protocols for opinion reconciliation}
	
	Cognitive maps and qualitative reasoning have been used as 
	modeling language for supporting decision in multi-agent context~\cite{chaib2002causal}.
	The concepts on which an agreement or disagreement exists are identified. 
	We rely on fuzzy logics to assess the degree of agreement, disagreement or conflict between two medical experts. %Being in a correlation with the real world, our system can not ignore the impact of relational uncertainty.
	Instead of concepts, we used natural language hypotheses generated from a medical ontology. 
	Based on personal construct theory, the approach in~\cite{chaib2002causal} focuses on understanding the perspective of the other agent on the given issue.
	Our assumption was that medical agents have constructed their opinions following an individual learning mechanism. 
	Hence, we are facing rather a black-box model than a transparent, explicit cognitive map, as assumed in~\cite{chaib2002causal}.
	Consequently, we designed argumentation protocols for each situation encountered, aiming to facilitate opinion reconciliation. 
	Degree of a relation (conflict, disagreement or agreement) can be used to assess the conflict dynamics.
	After each argumentation move, one disagreement situation transforms into another one with a different fuzzy degree of conflict. 
	Hence, argumentation protocols can be seen as conflict resolution strategy with its effectiveness quantified by the current degree of conflict. 
	
	Several collaborative knowledge models integrate argumentation protocols with machine learning,  such as Delphi technique~\cite{gobbi2012expert} and the knowledge spiral model~\cite{xiong2012argumentation}.
	The Delphi technique is a structured, iterative, anonymous collaborative method based on three phases for structuring the communication process: exploration, understanding, and evaluation. 
        The technique is the most effective in situations in which there is a strong degree of conflict among parties~\cite{gobbi2012expert}. 
	The knowledge spiral model~\cite{xiong2012argumentation} is composed by four modes of knowledge conversion: socialization, externalization, combination and internalization, that are considered the main stages of knowledge creation within an organization. 
	The process starts from the individual, the initial standalone information being integrated in explicit concepts by sharing and combining them. 
	This concept is used also in our approach, implying that jointly learning from argumentation can achieve the success of knowledge-creation.
	The combination of knowledge, dialog and reasoning individuals is also discussed in ~\cite{Letia2012}.  
	%The argumentation process between agents through distributed justification logic (DJL). DJL  is part of epistemic logics, developing the idea of reasoning based on evidence.
	
	 The problem of quantifying disagreement in argumentation systems is also addressed in~\cite{booth2012quantifying}. Here, the solution implies the measurement of distance between labellings of a graph representation. For achieving the reconciliation proposed, our system relies on a fuzzy quantification of arguments.
	In line with~\cite{hao2014arguing} we exploit argumentative protocols for opinion reconciliation between different learners. While in~\cite{hao2014arguing} the classification of data is based on inducing modular rules by Prism algorithms, our approach focuses on the correlation between fuzzy rules and textual entailment results.

	%\paragraph{Arguing in medical domain}
       %\subsection{Designining the multi-agent harmonisation protocols}
	%Argumentation is based on the reasoning dialog processes which, given several opinions, have as objective the inference of a conclusion. In the view of multi-agent systems, argumentation is correlated with an exchange of information between a set of agents, in order to resolve a given problem. The agents construct, analyze and evaluate the arguments and finally draw a justified conclusion.
	
	The dialog between two particular agents is considered a process of learning from experience~\cite{wardeh2012multi}. Joint learning in a conflict argumentation environment improves the learning ability of each individual agent, as a reference to the model of argument games~\cite{yao2012evaluating}.
	
	Speech acts in argumentative discourses~\cite{van1984speech} have an important role in resolving a conflict. In multi-agent environments, the message passing between agents is controlled by protocols. Complex conversations are governed by well defined interaction protocols. The mediation dialogs are created by the execution of these protocols. The agents must have the ability to observe, manipulate, reason, challenge and deliberate.

	The value of argumentation in reaching agreements derives from its capability to deal with conflicts and uncertainty~\cite{modgil2013added}.
	Conflicting knowledge and opinions do occur between physicians~\cite{craven2012efficient}, given different experiences and knowledge sources.
	The decision-making systems are trying to find the best method of measuring and eliminating the conflict factor. 
	The methodology for designing medical debates in~\cite{craven2012efficient} relies on assumption-based argumentation (ABA). 
	In ABA, arguments are assessed from logic rules and the situation of attack is defined as a contradictory information to the assumption made.  Defeasible logic is used to model that preferences should be on a higher level than rules.
%	One of the most developed approaches in the application of Description Logics to medical terminologies is provided by OpenGALEN~\cite{rector2003opengalen}. 
	
	%\paragraph{Textual entailment and argumentation}

	\section{Conclusions}
	\label{sec:conclusion}
	A solution for the mediation process of conflicting information in the domain of Parkinson's disease is proposed. 
	Our approach is exploring textual entailment achievements in combination with joint learning. A multi-agent environment was considered for this purpose and fuzzy logic was used for the quantification of opinions and relation between them. The advantage of our system is that the study case can be easily varied. However, a drawback is considered to be the necessity of building big data sets in order to obtain higher accuracy.   
	
	Our approach may be also relevant in domains such as i) ontology update or ii)  argumentation mining. 
	Firstly, the system could be used as a method for validating or updating medical ontology axioms against new findings described by medical research.
	Secondly, the proposed approach  represents a new method to extract arguments from natural language texts, method that is based on textual entailment. 
	%\paragraph{Ensemble learning} It is also a step forward in the research domain of methods based on argumentation to aggregate opinions of different base learners.
	
	%\paragraph{Argumentation mining} 
	The following  extensions of our work can also be considered for further investigation such as:
	i) solving conflicts among more than two agents, scaling the size of interactions;
ii) grouping several agents with the same opinion and extending the dialog into a debate.
		%\item Automating the process of data set building
	
	% \fi
	\IEEEpeerreviewmaketitle

	%\section*{Acknowledgements}

	\bibliographystyle{IEEEtran}      % mathematics and physical sciences
	\bibliography{bib}

\end{document}